\documentclass[a4paper, 10pt, conference]{ieeeconf}      
\usepackage{FG2027}
\usepackage{graphicx}
\usepackage{xcolor}   
\usepackage{amssymb}  
\usepackage{cite}
\usepackage[hidelinks]{hyperref}
\usepackage{tikz}
\usepackage{booktabs}
\usetikzlibrary{arrows.meta,fit,backgrounds,calc}

\FGfinalcopy 

\IEEEoverridecommandlockouts                              
\def\FGPaperID{350} 

\title{\LARGE \bf
Egocentric Action Recognition with Objects and Vision-Language Affordances
}

\author{\parbox{16cm}{\centering
    {\large Ra\'ul Boix-Aldeguer$^1$, Manuel Benavent-Lledo$^{2}$, David Ortiz-Perez$^1$ and Jose Garcia-Rodriguez$^1$}\\
    {\normalsize
    $^1$ Department of Computer Technology, University of Alicante, Alicante, Spain\\
    $^2$ Visual Analysis and Perception Lab, Aalborg University, Denmark}}
}

\begin{document}

\ifFGfinal
\thispagestyle{empty}
\pagestyle{empty}
\else
\author{Anonymous FG2027 submission\\ Paper ID \FGPaperID \\}
\pagestyle{plain}
\fi
\maketitle

\begin{abstract}

Action recognition, and more specifically egocentric action recognition, is a core task in computer vision, with applications such as monitoring the daily life activities of elderly people and task assistance. Recent methods rely on visual modalities such as RGB video or depth. However, in egocentric video, an action is largely defined by its objects, which these methods rarely model. We propose a method that combines a frozen SlowFast video expert with an attention-based object expert by late fusion. Each object is represented with its affordances, that is, what can be done with it, generated zero-shot by a vision-language model. Since datasets such as MECCANO annotate only the objects involved in each action, evaluating with these boxes tells the model which objects are active. We therefore train on annotated boxes and evaluate the method with detections from the official active-object detector, and with annotated boxes as an upper bound. Without affordances, our fusion reaches 68.4\% Top-1 with annotated boxes but 49.0\% with detections. About half of this drop comes from non-active detections. To remove them, the same vision-language model generates an interaction affordance that decides whether a hand is manipulating each detected object. On MECCANO, objects raise the Top-1 of the video expert from 44.9\% to 49.0\%, and the two affordances raise it to 49.9\%, above SlowFast and object-aware RGB baselines. Code will be released upon acceptance.

\end{abstract}

\section{INTRODUCTION}

Egocentric action recognition identifies the actions performed by a person from videos captured from their own point of view. It has applications in robotics, augmented reality, and personalized assistance, such as monitoring the daily activities of elderly people or guiding a user through a task. The spread of smart glasses and head-mounted cameras has generated a growing interest in understanding human activities from a first-person perspective, and several egocentric datasets support its study~\cite{li2018eye,damen2022rescaling,sener2022assembly101,ragusa2021meccano,grauman2022ego4d}. 

\begin{figure}[t]
    \centering
    \includegraphics[width=\columnwidth]{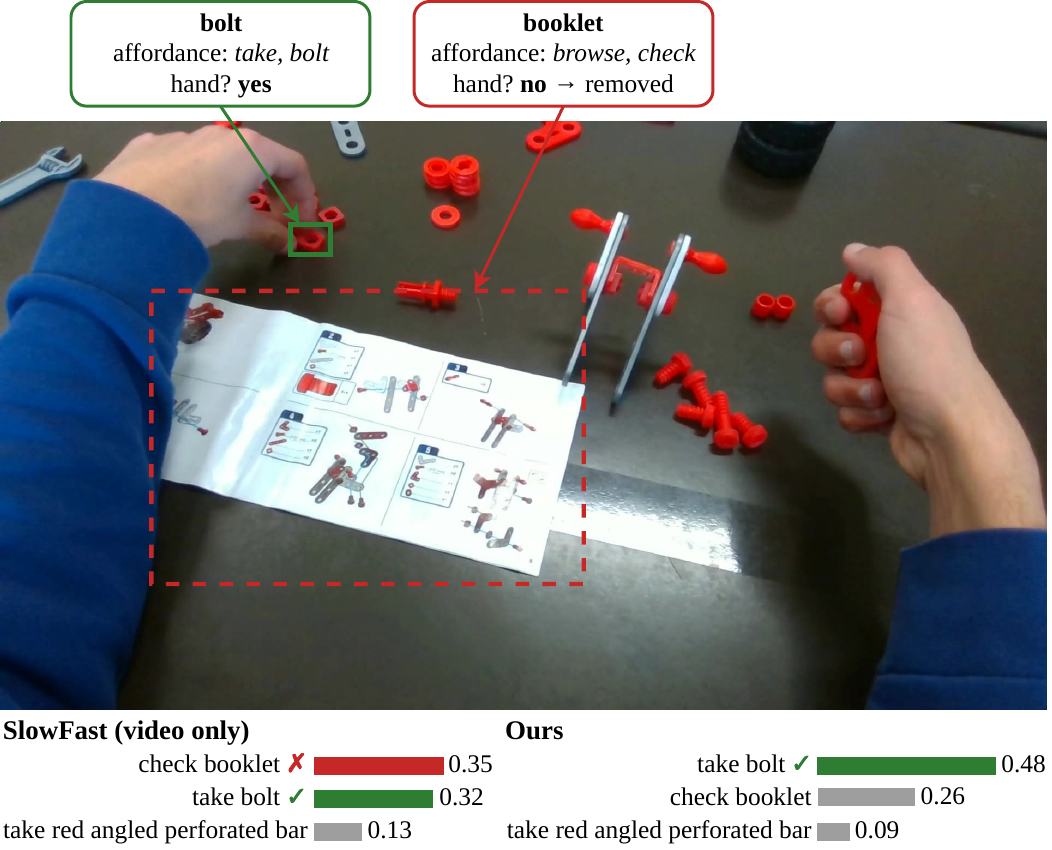}
    \caption{The interaction affordance removes an idle object. The VLM answers that no hand manipulates the booklet (red, dashed), and the filter removes it. The video-only SlowFast baseline and the fusion with all detected objects predict \emph{check booklet}, whereas our method predicts the correct action, \emph{take bolt}. Bars show the three most probable actions, averaged over five seeds for our method.}
    \label{fig:teaser}
\end{figure}

Most action recognition methods learn holistic spatio-temporal features from the whole clip, with 3D convolutional networks~\cite{tran2015learning,feichtenhofer2019slowfast} or video transformers~\cite{bertasius2021space,tong2022videomae}. Egocentric methods add further modalities such as depth, gaze, or audio~\cite{ragusa2023meccano,kazakos2019epic}. These models do not represent the objects explicitly, although in egocentric video the action is largely determined by the objects and what can be done with them. For example, screwing a screw by hand and with a screwdriver involve similar hand motions, and only the manipulated screwdriver tells them apart. The few recent egocentric methods that use objects add them as extra supervision during training~\cite{shiota2024egocentric} or combine detected regions with gaze~\cite{haralovic2026g3ego}, but they do not model what the objects afford.

A second limitation is how object-based models are evaluated. They are usually tested with annotated boxes, but egocentric datasets such as MECCANO~\cite{ragusa2021meccano,ragusa2023meccano} annotate only the active objects, so these boxes tell the model which objects take part in the action. However, a detector also reports objects that are present but not manipulated, so the accuracy drops. Works that replace annotations with hand-object~\cite{shan2020understanding} or open-vocabulary detectors~\cite{haralovic2026g3ego} rarely measure how much accuracy is lost with respect to annotated boxes, or why.

In contrast, we propose a method that combines a frozen SlowFast video expert with an attention-based object expert by late fusion. Each object is represented by its appearance, class name, position, and affordances, which describe what can be done with the object in that situation~\cite{gibson1979ecological}. Our hypothesis is that a zero-shot vision-language model (VLM) provides object knowledge that the video expert does not learn from the training videos. The VLM generates the affordances of each object as a short text. It also decides whether a hand is manipulating each detected object, and this interaction affordance removes idle detections before they reach the object expert. Fig.~\ref{fig:teaser} shows an example in which the filter removes an idle instruction booklet, and the method recovers the correct action. We evaluate the method on MECCANO using the official active-object detector and annotated boxes as an upper bound. With detections, it improves the video expert by 4.9 Top-1 and 8.4 Top-5 points and outperforms SlowFast and object-aware RGB baselines. Our contributions can be summarized as follows:

\begin{itemize}
  \item A method for egocentric action recognition that fuses a frozen video expert with an attention-based object expert, including affordances generated zero-shot by a VLM.
  \item An interaction affordance, generated by the same VLM, that removes, at test time, detected objects not manipulated at that moment, so fewer non-active objects reach the object expert.
  \item An analysis of the drop from annotated boxes to detections, which shows that 48\% of it comes from non-active detections.
\end{itemize}

\section{RELATED WORK}

\textbf{Egocentric action recognition.}
Egocentric action recognition methods usually build on video backbones such as SlowFast~\cite{feichtenhofer2019slowfast} and add signals from wearable cameras. For instance, EPIC-Fusion~\cite{kazakos2019epic} combines RGB, optical flow, and audio, and EgoVLP~\cite{lin2022egocentric} pretrains video and language representations on egocentric clips. On MECCANO, the dataset authors report SlowFast baselines with RGB, depth, and gaze, and combining the three modalities improves over RGB alone. A late fusion of RGB and depth Swin3D models won first place in the MECCANO action recognition challenge~\cite{kini2023egocentric}. More recently, G3Ego~\cite{haralovic2026g3ego} builds graphs from gaze, hands, and objects. An open-vocabulary detector localizes these objects from the captions of a VLM, and gaze discards the objects that the camera wearer does not look at. In these methods, however, objects are not described by the affordances they offer. Like G3Ego, our method excludes irrelevant objects, but uses a VLM-based manipulation filter rather than gaze.

\textbf{Object-based action recognition.}
Objects have been a central cue in egocentric video since early works. Fathi et al.~\cite{fathi2011learning} learn object models from egocentric activities with weak supervision, and Pirsiavash and Ramanan~\cite{pirsiavash2012detecting} observe that objects look different while they are being used and train separate detectors for active and passive objects. More recent models reason over the boxes of hands and objects. STIN~\cite{materzynska2020something} models how hand and object boxes interact over time and reports results with both annotated and detected boxes. Shiota et al.~\cite{shiota2024egocentric} use hand-object contact and object state as extra supervision during training, so inference needs only RGB video. Hand-object detectors~\cite{shan2020understanding} localize hands and the objects they touch, and MECCANO's active-object detector localizes the objects involved in the action. However, a detector also reports objects that are present but not used, and annotated boxes hide this problem because they mark only the active objects. Our object expert also uses boxes, but it represents each object with its affordances and removes the detections that are not being manipulated.

\textbf{Affordances in egocentric video.}
Affordances describe what an agent can do with an object~\cite{gibson1979ecological}. Nagarajan et al.~\cite{nagarajan2019grounded} learn interaction hotspots, maps of where an object is likely to be manipulated, by watching people use objects. Ego-Topo~\cite{nagarajan2020ego} links environmental zones to the activities they afford to anticipate future actions. VRB~\cite{bahl2023affordances} learns from egocentric videos where and how people interact with a scene and transfers these affordances to robots. AFF-ttention!~\cite{murlabadia2024aff} introduces STAformer, which uses environment affordances and interaction hotspots to anticipate the next object interaction in egocentric video. These works localize or predict affordances to anticipate interactions or guide robots, whereas our method uses short text descriptions of each object for action recognition.

\begin{figure*}[t]
  \centering
  \includegraphics[width=\textwidth]{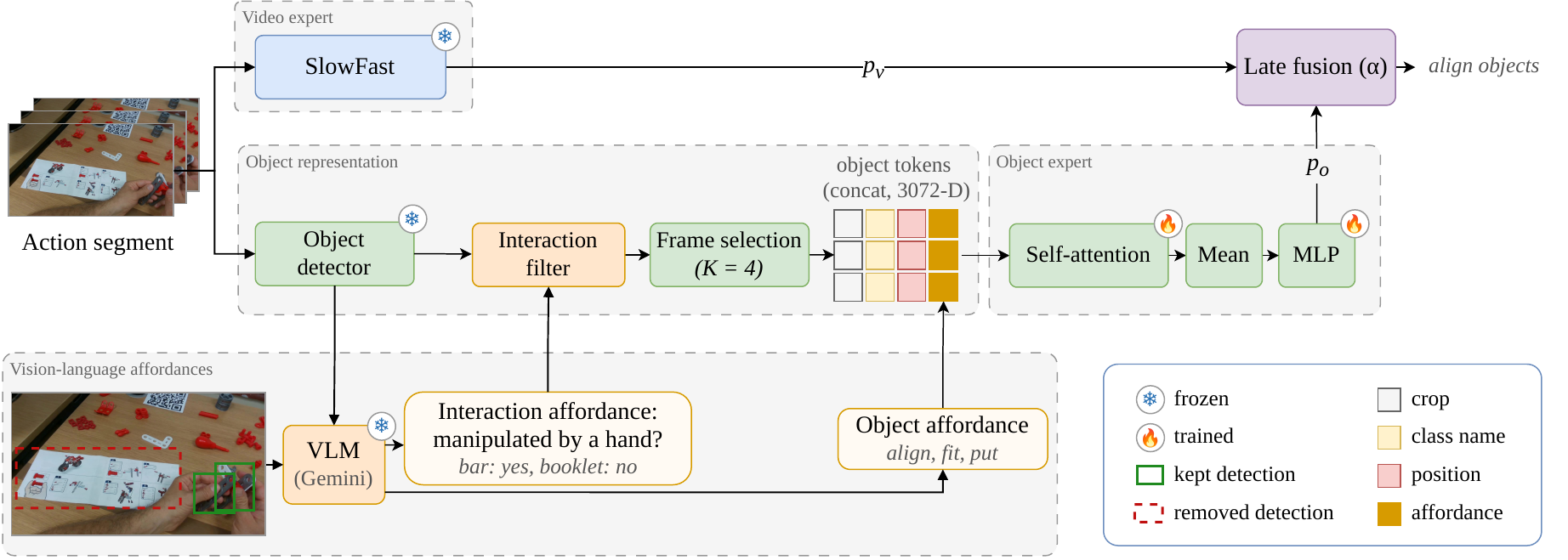}
  \caption{Overview of the method at test time. A frozen SlowFast expert gives the video probabilities $p_v$. An object detector finds the objects in the candidate frames of the segment. For each detection, a VLM answers whether a hand is manipulating the object, and the interaction filter removes the detections answered with \textit{no} (red, dashed). Then $K$ frames are selected, and each remaining object becomes a token of crop, class name, position, and affordance features. The affordance is generated by a separate prompt to the same VLM. The object expert applies self-attention to the tokens, averages them, and an MLP gives the probabilities $p_o$. Finally, late fusion combines both experts with the weight $\alpha$. During training, the annotated boxes replace the detector, and the filter is not used.}
  \label{fig:method}
\end{figure*}

\textbf{Language for action recognition.}
Language also provides knowledge about actions. ActionCLIP~\cite{wang2025actionclip} treats action recognition as matching videos with the text of the class names, and LaViLa~\cite{zhao2023learning} adapts a large language model to narrate egocentric videos and trains video-language representations on these narrations. These methods describe whole clips or classes, whereas our method describes each detected object and what it affords.

\textbf{Vision-language models for affordances.}
Vision-language models learn from large image-text corpora that often describe how objects are used, which makes them a source of affordances. AffordanceLLM~\cite{qian2024affordancellm} fine-tunes a VLM to ground affordance regions in images, and SpotLight~\cite{engelbracht2025spotlight} uses VLM affordance predictions to select robot motion primitives. Tonini et al.~\cite{tonini2026towards} evaluate multimodal LLMs that describe human-object interactions in free-form text and compare visual prompts such as red circles and crops. Unlike these works, we use a VLM zero-shot, first, to describe each detected object and its affordances for action recognition, and second, to remove detections that a hand is not manipulating.

\section{METHOD}\label{sec:method}

Given a video segment that contains one action, the method predicts its class among $C$ action classes. The input is the RGB segment and the objects in a set of candidate frames of the segment, each one a box with an object class. During training, the boxes come from annotations of the objects involved in the action (active objects). At test time, an object detector trained on the same object classes gives the boxes, and it also finds objects that are present but not involved in the action (non-active objects). We do not train the method on detections, because the detector is trained on the same training videos and is almost perfect on them, so it does not make the errors it makes on new videos.

Fig.~\ref{fig:method} shows the pipeline at test time. A frozen video expert computes class probabilities from the whole segment (Sec.~\ref{sec:experts}). In parallel, the object detector finds objects in the candidate frames, and a VLM generates two affordances for each detection (Sec.~\ref{sec:affordances}): an object affordance describing what can be done with the object, and an interaction affordance that removes the objects that are not manipulated. The object expert classifies the remaining objects (Sec.~\ref{sec:experts}), and late fusion combines both experts (Sec.~\ref{sec:late}). The interaction filter is used only at test time, because all annotated boxes are active objects.

\subsection{Video and Object Experts}\label{sec:experts}
    \subsubsection{Video expert}

    The video expert is a video classifier trained on the target dataset and kept frozen. We use SlowFast~\cite{feichtenhofer2019slowfast}, which samples frames over the whole segment and averages the class probabilities $p_v$ of several spatial crops.

    \subsubsection{Object representation}

    The object expert observes $K=4$ frames per segment. Among the candidate frames of the segment, it selects the $K$ frames with the largest change in the set of object classes with respect to the previous candidate frame (maximize transitions). Frames where the set of objects changes tend to mark the steps of the action, such as taking a part or aligning two parts. If a segment has fewer than $K$ candidate frames, all of them are used. During training, the boxes and classes come from the annotations. At test time, the selection uses the classes of the filtered detections (Sec.~\ref{sec:affordances}).

    Each object $i$ of the $N$ objects in the $K$ selected frames is represented by four parts: visual appearance, class name, position, and affordance. The visual appearance $\mathbf{v}_i$ is the [CLS] embedding of a ViT-B/16~\cite{dosovitskiy2021image}, pretrained on ImageNet-21k and fine-tuned on ImageNet-1k, applied to the box crop. The class name, a text with the box corners, size, and center normalized by the frame size, and the affordance text generated by the VLM (Sec.~\ref{sec:affordances}) are encoded with DistilBERT~\cite{sanh2019distilbert} ([CLS] token), giving the 768-dimensional embeddings $\mathbf{c}_i$, $\mathbf{b}_i$, and $\mathbf{a}_i$. Representing the affordances as text keeps the VLM frozen and lets the same text encoder process the class name, the position, and the affordances. The object token is
    \begin{equation}
    \mathbf{x}_i = \left[\mathbf{v}_i;\, \mathbf{c}_i;\, \mathbf{b}_i;\, \phi(\mathbf{a}_i)\right] \in \mathbb{R}^{3072},
    \end{equation}
    where $\phi$ is a trained projection from 768 to 768 dimensions (linear layer, layer normalization, and ReLU) and all encoders are frozen.

    \subsubsection{Object expert}

    The object expert projects each token to $\mathbf{h}_i \in \mathbb{R}^{512}$ (linear layer, layer normalization, and ReLU), applies self-attention among the objects, and averages the result:
    \begin{equation}
    H' = \mathrm{LN}\big(H + \mathrm{MHA}(H, H, H)\big), \qquad
    \mathbf{o} = \frac{1}{N} \sum_{i=1}^{N} \mathbf{h}'_i,
    \end{equation}
    where $H = [\mathbf{h}_1, \dots, \mathbf{h}_N] \in \mathbb{R}^{N \times 512}$ stacks the $N$ object tokens, MHA is multi-head self-attention~\cite{vaswani2017attention} with four heads, LN is layer normalization, $\mathbf{h}'_i$ is the $i$-th row of $H'$, and $\mathbf{o}$ is a single representation of the whole segment. A segment without objects ($N=0$) uses a learned token instead. An MLP with two hidden layers (512 and 256 units) maps $\mathbf{o}$ to the logits $\mathbf{z}_o$, and the expert is trained with cross-entropy.

    Only $\phi$, the token projection, the self-attention layer, the token for segments without objects, and the MLP are trained, while the video expert, ViT, DistilBERT, the detector, and the VLM remain frozen.

\subsection{Vision-Language Affordances}\label{sec:affordances}
    \subsubsection{Object affordances}

\begin{figure*}[t]
      \centering
      \includegraphics[width=\textwidth]{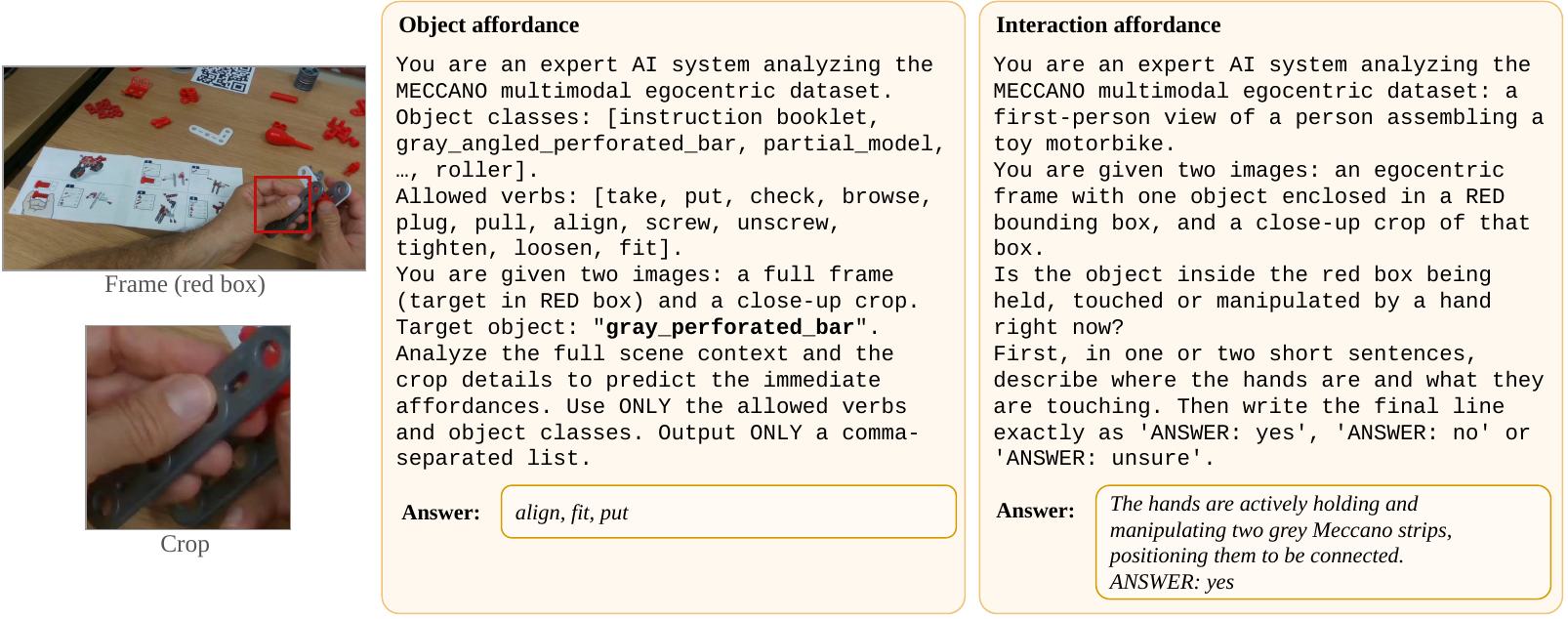}
      \caption{Prompts of the two VLM queries for the gray bar of Fig.~\ref{fig:method}. Both queries receive the frame with the detection in a red box and its crop. The object affordance (center) returns a comma-separated list, which is encoded with DistilBERT. The interaction affordance (right) describes the hands and answers whether a hand manipulates the object; detections answered \emph{no} are removed. Prompts used on MECCANO. The list of object classes is shortened.}
      \label{fig:prompt}
    \end{figure*}

    A VLM generates the affordances of each object zero-shot as a short text. The VLM receives the class name, the full frame with the object in a red box, and the crop of the object. The red box tells the VLM which object to describe while preserving scene context, and the crop shows the object at higher resolution, as in recent work comparing visual prompts for describing interactions~\cite{tonini2026towards}. The prompt describes the task and restricts the answer to a comma-separated list built from the verb and object classes of the dataset (Fig.~\ref{fig:prompt}, center).

    \subsubsection{Interaction affordance}

    The annotated boxes used for training contain only active objects, but the detector also reports objects that are present but not active. To address this, the method asks the VLM whether a hand is manipulating each detected object at that moment. We call this the interaction affordance. If the answer is \textit{no}, the object is removed. For each detection $d$, the VLM receives the frame with the object in a red box and its crop. It first describes where the hands are, and then answers whether a hand holds, touches, or manipulates the object (\textit{yes}, \textit{no}, or \textit{unsure}) (Fig.~\ref{fig:prompt}, right). Describing the hands first is meant to make the VLM look at the hands and the object together, instead of answering from the object alone.

    Let $\mathcal{D}_t$ be the detections in frame $t$, $s(d)$ the detector score, and $r(d)$ the VLM's answer. The filter keeps
    \begin{equation}
    \mathcal{D}'_t = \{\, d \in \mathcal{D}_t : r(d) \neq \textit{no} \,\} \cup \{\, \arg\max_{d \in \mathcal{D}_t} s(d) \,\},
    \end{equation}
    for every frame with at least one detection, so each such frame keeps its most confident detection. Detections answered \textit{unsure} are kept. The filter runs on detections before frame selection.

\subsection{Late Fusion}\label{sec:late}

The final prediction combines both experts:
\begin{equation}
\hat{y} = \arg\max_{k} \big[\, \alpha \log p_v(k) + (1-\alpha) \log p_o(k) \,\big],
\end{equation}
where $k$ ranges over the $C$ action classes, $p_o = \mathrm{softmax}(\mathbf{z}_o)$, and $\alpha \in [0, 1]$ is chosen on the validation set with a step of 0.05. With $\alpha = 1$, the method uses only the video expert, and with $\alpha = 0$, only the object expert.

\section{EXPERIMENTS}

\subsection{Experimental Setup}\label{sec:exp_setup}

\textbf{Dataset.} MECCANO~\cite{ragusa2021meccano,ragusa2023meccano} contains 20 egocentric videos of people assembling a toy motorbike, annotated with 61 action classes, which combine  12 verbs and 20 object classes. The official split has 11, 2, and 7 videos for training, validation, and test, with 5,057, 977, and 2,823 action segments. Three exact duplicate rows are removed from the test annotations, which leaves 2,820 test segments. Only the objects involved in the action are annotated with boxes, and the active-object annotations include only frames with at least one active object. We use MECCANO because it annotates the active objects of every action segment with boxes and provides an official active-object detector for the same classes. Both are needed to train the object expert and to measure the drop from annotated boxes to detections.

\textbf{Evaluation protocol.} The method is evaluated in two settings. With \emph{annotated boxes} as an oracle, the object expert uses the annotations for training, validation, and testing, providing an upper bound. With \emph{detections}, the object expert is trained on annotated boxes (Sec.~\ref{sec:method}), and the validation and test sets use the official active-object detector of MECCANO with a score threshold of 0.5. The interaction filter is applied to the validation and test detections, and $\alpha$ and early stopping use the validation set with detections.

\textbf{Detector.} The detector is a Faster R-CNN~\cite{ren2015faster} and it runs on the frames listed in the active-object annotations (the candidate frames), without their boxes. On the test set, it reaches 31.6 AP at IoU 0.5, close to the 30.39 reported by the MECCANO authors~\cite{ragusa2021meccano}. A detection is \emph{active} if it matches an annotated box with an IoU of at least 0.5, and \emph{non-active} otherwise. Of the detections, 53.9\% are non-active, and 62.0\% of the annotated boxes are found. The AP is low partly because only active objects are annotated, so correct boxes on non-active objects count as errors. On the training set, it reaches 94.3 AP, so the object expert is not trained on detections.

\textbf{Model variants.} The \emph{video expert} is SlowFast alone. \emph{Video + objects} is the late fusion of the video expert and the object expert, without affordances or interaction filter. \emph{Video + objects + VLM} is our method, which adds the affordances and the interaction filter. Unless stated otherwise, results use detections.

\textbf{Metrics.} The metrics are Top-1 and Top-5 accuracy and macro-F1. We include macro-F1 to show the performance on rare actions, as the classes are imbalanced (the most frequent action covers 27.3\% of the test segments). Results are the mean of five seeds, with the standard deviation for Top-1. Differences between two models are reported with a 95\% confidence interval from a paired bootstrap over the test segments. All results are on the test set unless stated otherwise, and * marks differences whose 95\% confidence interval excludes zero.

\textbf{Implementation details.} The video expert is SlowFast R50 ($8\times8$) with the checkpoint released with MECCANO. It samples 32 frames over the segment and averages three spatial crops of $256 \times 256$ that cover the full frame width, as in the official test protocol. The object expert is trained with Adam (learning rate $2\cdot10^{-4}$, no weight decay, batch size 64, dropout 0.5) for at most 100 epochs, with early stopping on the validation Top-1 (patience of 15 epochs). The VLM is Gemini 3.1 Flash-Lite~\cite{gemini} with temperature 0. It runs offline in the batch mode of Vertex AI, and each detection needs two queries, one for its affordance and one for the interaction affordance. On the test set, this amounts to 53,748 queries. Each query has 2,300 to 2,400 input tokens, and each test segment needs about 19 queries.

\begin{table}[t]
  \caption{Comparison with published results on MECCANO. \emph{Extra input}: what each method uses at inference besides RGB video.}
  \label{tab:baselines}
  \centering
  \footnotesize
  \setlength{\tabcolsep}{2.5pt}
  \begin{tabular}{@{}llccc@{}}
    \toprule
    Method & Extra input & Top-1 & Top-5 & Macro-F1 \\
    \midrule
    SlowFast~\cite{ragusa2023meccano} & -- & 45.16 & 73.75 & 18.91\rlap{$^\dagger$} \\
    HOCL+OSL~\cite{shiota2024egocentric} & -- & 44.81 & 77.01 & 16.70 \\
    UCF Swin3D-B~\cite{kini2023egocentric}$^\ddagger$ & -- & \textbf{50.80} & 82.36 & -- \\
    SlowFast~\cite{ragusa2023meccano} & D+G & 49.66 & 77.82 & -- \\
    G3Ego~\cite{haralovic2026g3ego} & G+H+VLM+OD & 46.48 & 82.04 & 21.34 \\
    \midrule
    Video expert & -- & 44.93 & 74.18 & 19.43 \\
    \quad + objects & AOD & 49.04\,{\scriptsize$\pm$0.68} & 81.34 & 22.69 \\
    \quad + objects + VLM (ours) & AOD+VLM & 49.87\,{\scriptsize$\pm$0.47} & \textbf{82.54} & \textbf{24.06} \\
    \midrule
    Video + objects & GT boxes & 68.38\,{\scriptsize$\pm$1.42} & 95.34 & 42.57 \\
    \bottomrule
  \end{tabular}
    \\[2pt]
    \parbox{\columnwidth}{\scriptsize D: depth; G: gaze; H: hands; OD: open-vocabulary detector; AOD: active-object detector; GT boxes: annotated boxes as an oracle. $^\dagger$Computed by~\cite{haralovic2026g3ego}. $^\ddagger$Pretrained on Something-Something v2, trained with focal loss.}
\end{table}

\subsection{Comparison with Baselines}\label{sec:exp_baselines}

Table~\ref{tab:baselines} compares the method with published results on the official test split. With detections, our method reaches 49.9\% Top-1 and 82.5\% Top-5. It improves over the SlowFast RGB baseline of MECCANO by 4.7 Top-1 and 8.8 Top-5 points, and over our video expert by 4.9 and 8.4 points. This video expert uses the SlowFast checkpoint released with MECCANO and stays within 0.5 points of its published Top-1 and Top-5. Our method also outperforms HOCL+OSL, which uses hand-object contact and object state as supervision during training, and G3Ego, which needs gaze at test time.

Its Top-1 is similar to that of SlowFast with RGB, depth, and gaze, and its Top-5 is 4.7 points higher. UCF Swin3D-B obtains a higher Top-1 (50.8\%) and a similar Top-5 (82.4\%) with RGB only, using a Swin3D-B backbone pretrained on Something-Something v2 and a focal loss. Among the methods without annotated boxes, ours has the highest Top-5 and macro-F1. Compared with the video expert, it improves macro-F1 by 24\% and Top-1 by 11\% (relative), suggesting that the objects help less frequent actions more than frequent ones. However, the comparison is not fully equivalent. The detector is trained on the boxes of the MECCANO training set and runs on the frames listed in the active-object annotations, which contain at least one active object. A deployed system would also process frames without active objects.

Using annotated boxes, Video + objects reaches 68.4\% Top-1, but this upper bound is not comparable to the other rows because the annotated boxes reveal the active objects. Sec.~\ref{sec:exp_drop} analyzes the gap.

\begin{figure}[t]
  \centering
  \includegraphics[width=\columnwidth]{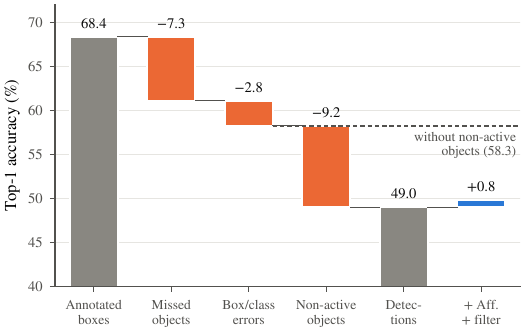}
  \caption{Decomposition of the Top-1 drop from annotated boxes to detections. Each orange bar is the loss of one step that replaces annotated boxes with detections. The blue bar is the gain of affordances and the interaction filter (Sec.~\ref{sec:exp_aff}). The dashed line marks the accuracy without non-active detections, the accuracy of a perfect interaction filter.}
  \label{fig:drop}
\end{figure}

\subsection{From Annotated Boxes to Detections}\label{sec:exp_drop}

Replacing the annotated boxes with detections lowers the Top-1 of Video + objects from 68.4\% to 49.0\%. To identify the causes of this drop, the annotated boxes of the test set are replaced by detections in three steps. First, the annotated boxes that the detector misses are removed. Second, the remaining annotated boxes are replaced by their matched detections, which adds errors in box position and class. Third, the non-active detections are added, which gives the detector output.

Fig.~\ref{fig:drop} shows that the non-active detections cause the largest loss, 9.2 points (48\% of the drop). A likely reason is that the object expert is trained on annotated boxes, in which every object takes part in the action. However, at test time, it sees every detected object, so it treats non-active objects as part of the action. Missed objects cost 7.3 points (38\%). The detector is weakest on small parts, such as the wrench, wheel axle, or rod, and strongest on large objects, such as the booklet and screwdriver.

Without non-active detections, Video + objects would reach 58.3\%. This is the accuracy of a perfect interaction filter, and it motivates the filter of Sec.~\ref{sec:affordances}. This estimate is optimistic, because some non-active detections are duplicates or badly placed boxes rather than idle objects. On the test set, 1,155 pairs of detections share the same box (IoU of at least 0.5) with different classes, up to 8\% of the non-active detections.

Table~\ref{tab:parts} shows why annotated boxes give an upper bound. With them, the class name alone is as accurate as the full token, and removing it costs 13.3 points. With detections, the class name alone loses 2.6 points, and the crop alone matches the full token. Annotated boxes therefore give the model the class of the active objects. Detections do not, and the appearance of the objects carries most of the information. The class name also loses the most from annotated boxes to detections (29.9 points, against 16.4 for the crop). With detections, the position alone falls below the most frequent action (26.1\% against 27.3\%), so the detected boxes carry little information without their appearance.

\begin{table}[t]
  \caption{Top-1 of the object expert alone with partial object tokens (mean pooling, no affordances).}
  \label{tab:parts}
  \centering
  \footnotesize
  \setlength{\tabcolsep}{4pt}
  \begin{tabular}{@{}lcc@{}}
    \toprule
    Token parts & Annotated boxes & Detections \\
    \midrule
    Full (crop + class + position) & 63.28 & 36.39 \\
    Class only & 63.71 & 33.83* \\
    Crop only & 52.81* & 36.39 \\
    Position only & 33.88* & 26.11* \\
    Crop + position (no class) & 50.00* & 35.27* \\
    \bottomrule
  \end{tabular}
\end{table}

\begin{table}[t]
  \caption{Ablation of the affordances and the interaction filter (late fusion, detections).}
  \label{tab:aff}
  \centering
  \footnotesize
  \setlength{\tabcolsep}{3pt}
  \begin{tabular}{@{}cccccc@{}}
    \toprule
    Aff. & Filter & Top-1 & Top-5 & Macro-F1 & $\Delta$ Top-1 \\
    \midrule
    -- & -- & 49.04\,{\scriptsize$\pm$0.68} & 81.34 & 22.69 & -- \\
    \checkmark & -- & 49.53\,{\scriptsize$\pm$0.49} & 82.08 & 23.36 & +0.50 {\scriptsize[+0.11, +0.93]}* \\
    -- & \checkmark & 49.54\,{\scriptsize$\pm$0.95} & 81.74 & 23.64 & +0.50 {\scriptsize[+0.04, +0.92]}* \\
    \checkmark & \checkmark & \textbf{49.87}\,{\scriptsize$\pm$0.47} & \textbf{82.54} & \textbf{24.06} & \textbf{+0.84} {\scriptsize[+0.30, +1.34]}* \\
    \bottomrule
  \end{tabular}
\end{table}

\begin{figure}[t]
  \centering
  \includegraphics[height=2.2cm]{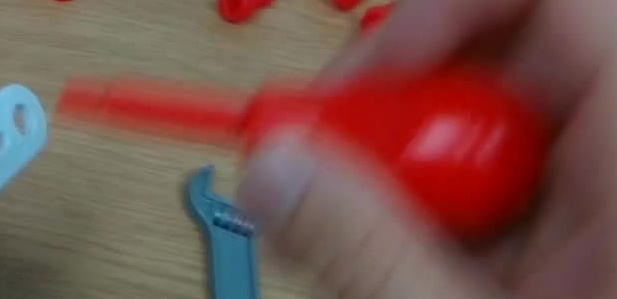}\hfill
  \includegraphics[height=2.2cm]{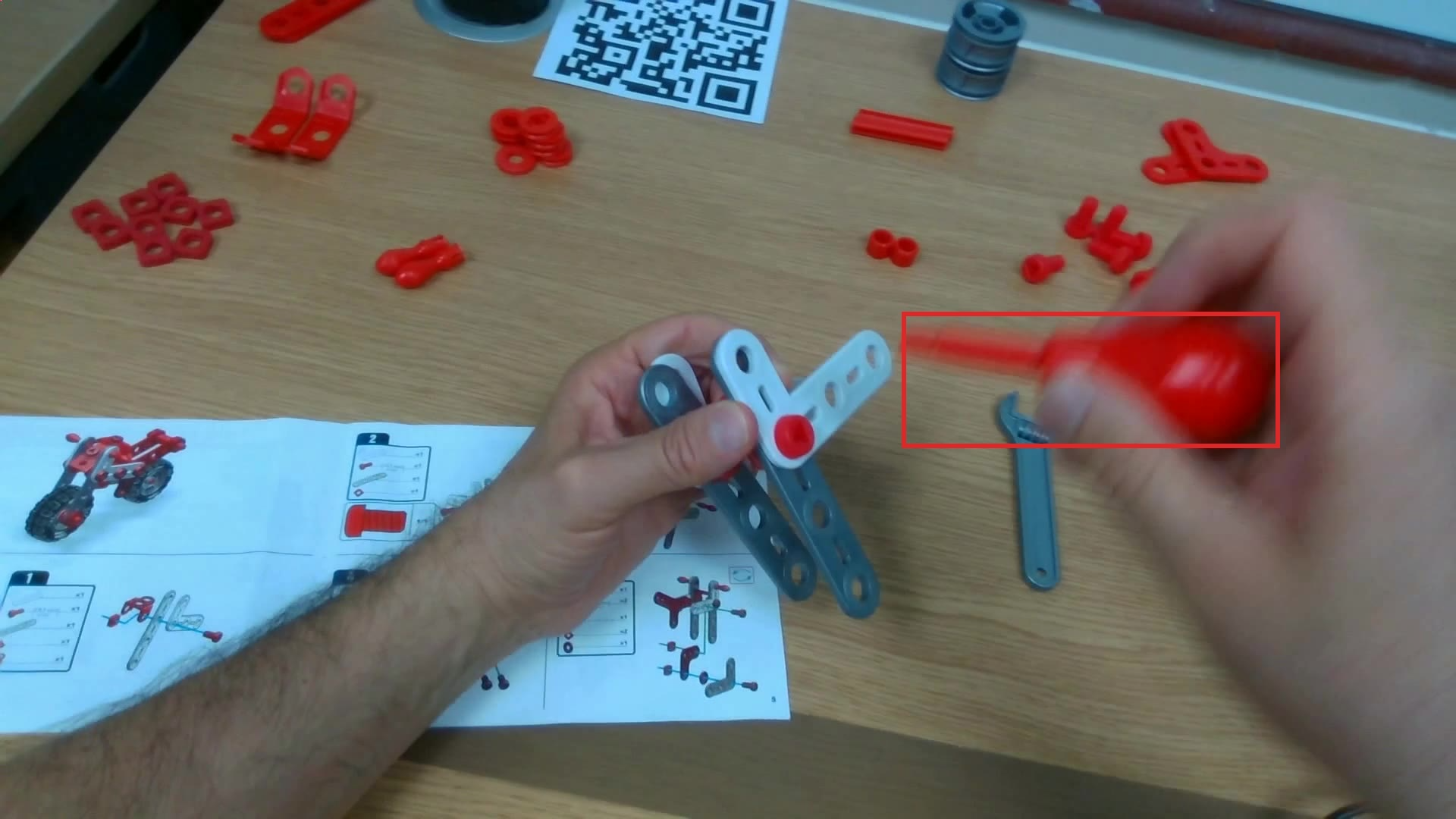}\\[3pt]
  {\scriptsize
  \begin{tabular}{@{}lp{0.66\columnwidth}@{}}
    Text only & take the screwdriver, align the screwdriver to the screw, tighten the screw, loosen the screw, put the screwdriver \\
    Text + crop & take screwdriver, screw screw, unscrew screw, tighten screw, loosen screw \\
    Text + frame & take, screwdriver \\
    Text + frame + crop & screw, tighten, unscrew, loosen \\
  \end{tabular}}
  \caption{Affordances generated by the VLM for a detected screwdriver with the four inputs: the crop (left) and the frame with the detection in a red box (right). The text-only answer depends only on the class name and is the same for every screwdriver. With the crop, the answer describes the screwdriver in use, whereas the frame alone gives a short answer.}
  \label{fig:affvariants}
\end{figure}

\subsection{Affordances and Interaction Filter}\label{sec:exp_aff}

Table~\ref{tab:aff} adds the affordances and the interaction filter to Video + objects with detections. $\Delta$ is the Top-1 difference to Video + objects, with its 95\% confidence interval. Each component improves Top-1 by 0.5 points. Together, they improve Top-1 by 0.8, Top-5 by 1.2, and macro-F1 by 1.4 points, all with intervals that exclude zero. Our method outperforms Video + objects in four of the five seeds. Across seeds, our method ranges from 49.4\% to 50.5\% Top-1, and Video + objects from 48.4\% to 50.2\%. The gains are small but consistent on the test set. The two components help in different ways: the affordances raise Top-5 more (0.7 against 0.4 points), and the filter raises macro-F1 more (1.0 against 0.7). Our method also has the lowest standard deviation between seeds (0.47, against 0.68 for Video + objects). On the validation set, which has only two videos, the differences are within the variation between seeds (52.6\% against 52.3\% for Video + objects). On the full validation set, the filter removes only 7.4\% of the non-active detections, which may explain why its gain does not appear there.

The prompt of the interaction affordance is chosen on 500 validation detections (Table~\ref{tab:intprompt}). For each prompt, the table gives the share of detections answered \emph{no} and the precision (P) and recall (R) of \emph{no} for non-active detections. It also gives the share of non-active and active detections removed by the filter (Eq.~3). Asking the VLM to describe the hands before answering removes the most non-active detections, with the same share of active ones as the frame and crop alone. On these detections, the VLM never answers \emph{unsure}, so the policy for this answer does not affect the results. On the test set, the interaction filter removes 12.1\% of the non-active detections and 3.7\% of the active ones, and the number of non-active detections per frame drops from 1.24 to 1.09. Keeping the most confident detection of each frame is important. Removing every detection answered \emph{no} (Table~\ref{tab:intprompt}, last row) would remove 20.5\% of the non-active detections, but also 18.6\% of the active ones. The VLM answers \emph{yes} for 79.5\% of the non-active detections, likely because many of them are close to the hands or even held (Fig.~\ref{fig:qual}c). In addition, an active object is not always touched, such as the booklet in a \emph{check booklet} action. As a result, the filter alone recovers 0.5 of the 9.2 points lost to non-active detections (Sec.~\ref{sec:exp_drop}).

We compare four variants that differ in the visual input: \textit{text only} (the class name, one query per class), \textit{text + crop}, \textit{text + frame} (the full frame with the object in a red box), and \textit{text + frame + crop}. The four affordance inputs improve Top-1 with detections by up to 0.5 points (Table~\ref{tab:affinputs}), and text~+~frame~+~crop, chosen on the validation set, gives the largest gain. Even text only, which gives the same affordances to every object of a class, improves Top-1 by 0.35 points with detections, so part of the gain comes from the knowledge of the VLM about each class rather than from the image. With annotated boxes, the affordances do not help consistently (between $-0.7$ and $+0.6$ points), because the annotated class already identifies the active objects (Table~\ref{tab:parts}). With detections, the VLM describes the object as it appears in the frame. In Fig.~\ref{fig:affvariants}, the text-only answer is the same for every screwdriver, but with the crop, the answer describes the screwdriver in use.

\begin{table}[t]
  \caption{Prompts of the interaction affordance.}
  \label{tab:intprompt}
  \centering
  \footnotesize
  \setlength{\tabcolsep}{3pt}
  \begin{tabular}{@{}lcccc@{}}
    \toprule
    VLM input & \emph{No} & P / R & \multicolumn{2}{c}{Removed} \\
    \cmidrule(l){4-5}
    & & & Non-active & Active \\
    \midrule
    \multicolumn{5}{@{}l}{\emph{Validation (500 detections)}} \\
    Frame & 11.0 & 60.0 / 14.1 & 8.1 & 2.3 \\
    + crop & 11.0 & 58.2 / 13.7 & 7.3 & 2.6 \\
    + hand description (used) & 14.2 & 56.3 / 17.1 & \textbf{9.4} & 2.6 \\
    \midrule
    \multicolumn{5}{@{}l}{\emph{Test (26,874 detections)}} \\
    + hand description (used) & 19.6 & 56.3 / 20.5 & 12.1 & 3.7 \\
    \quad without keeping the top detection & 19.6 & 56.3 / 20.5 & 20.5 & 18.6 \\
    \bottomrule
  \end{tabular}
\end{table}

\begin{table}[t]
  \caption{Top-1 difference of each affordance input to Video + objects, without interaction filter.}
  \label{tab:affinputs}
  \centering
  \footnotesize
  \begin{tabular}{@{}lcc@{}}
    \toprule
    VLM input & Annotated boxes & Detections \\
    \midrule
    Text only & $-$0.06 & +0.35* \\
    Text + crop & +0.55* & +0.41* \\
    Text + frame & $-$0.72* & +0.02 \\
    Text + frame + crop (used) & $-$0.54 & \textbf{+0.50*} \\
    \bottomrule
  \end{tabular}
\end{table}

\begin{table}[t]
  \caption{Fusion strategies without affordances.}
  \label{tab:fusion}
  \centering
  \footnotesize
  \setlength{\tabcolsep}{2.5pt}
  \begin{tabular}{@{}lcccc@{}}
    \toprule
    & \multicolumn{3}{c}{Detections} & Annotated \\
    \cmidrule(lr){2-4}\cmidrule(l){5-5}
    Model & Top-1 & Top-5 & Macro-F1 & Top-1 \\
    \midrule
    Video expert & 44.93 & 74.18 & 19.43 & 44.93 \\
    Object expert, mean & 36.39\,{\scriptsize$\pm$0.38} & 72.00 & 11.20 & 63.28\,{\scriptsize$\pm$0.33} \\
    Object expert, attn. & 37.75\,{\scriptsize$\pm$0.55} & 74.69 & 12.84 & 64.27\,{\scriptsize$\pm$0.70} \\
    \midrule
    Learned: sum, mean & 48.13\,{\scriptsize$\pm$0.58} & 80.19 & 21.64 & 58.44\,{\scriptsize$\pm$0.25} \\
    Learned: concat., mean & 47.29\,{\scriptsize$\pm$0.36} & 77.03 & 21.02 & 55.26\,{\scriptsize$\pm$0.32} \\
    Learned: sum, attn. & 48.31\,{\scriptsize$\pm$0.31} & 79.33 & 22.16 & 57.97\,{\scriptsize$\pm$0.33} \\
    Learned: weighted sum, attn. & 48.10\,{\scriptsize$\pm$0.38} & 78.77 & 22.07 & 57.72\,{\scriptsize$\pm$0.38} \\
    Learned: cross-attention & 46.86\,{\scriptsize$\pm$0.70} & 74.95 & 21.06 & 54.91\,{\scriptsize$\pm$0.42} \\
    \midrule
    Late, mean & 48.62\,{\scriptsize$\pm$0.32} & 80.74 & 22.45 & 68.25\,{\scriptsize$\pm$0.48} \\
    Late, attn. (used) & \textbf{49.04}\,{\scriptsize$\pm$0.68} & \textbf{81.34} & \textbf{22.69} & \textbf{68.38}\,{\scriptsize$\pm$1.42} \\
    \bottomrule
  \end{tabular}
\end{table}

\begin{table}[t]
  \caption{Frame selection and number of frames $K$ (late fusion, no affordances).}
  \label{tab:frames}
  \centering
  \footnotesize
  \setlength{\tabcolsep}{3pt}
  \begin{tabular}{@{}lccc@{}}
    \toprule
    Frames & $K$ & Detections & Annotated \\
    \midrule
    Maximize transitions (used) & 4 & 49.04 & 68.38 \\
    Fixed offsets & 4 & $-$0.45 & $-$4.55* \\
    Uniform annotated & 4 & \textbf{+0.72*} & +2.05* \\
    High variability & 4 & $-$0.06 & $-$2.98* \\
    Progressive discovery & 4 & +0.33 & $-$1.01* \\
    \midrule
    Maximize transitions & 6 & +0.22 & +2.06* \\
    Uniform annotated & 6 & +0.20 & \textbf{+2.92*} \\
    Uniform annotated & all & $-$0.01 & +1.84* \\
    \bottomrule
  \end{tabular}
\end{table}

\subsection{Fusion and Frame Selection}\label{sec:exp_fusion}

This section justifies two design choices of Sec.~\ref{sec:method}, the fusion strategy and the frame selection, both evaluated without affordances.

Table~\ref{tab:fusion} compares the fusion strategies. The learned fusions combine the video embedding and the pooled object representation before a shared MLP, and the object tokens are pooled with mean or attention (attn.). With annotated boxes, late fusion outperforms all learned fusions by about 10 points, and the learned fusions are below the object expert alone. The two experts are complementary. For example, with detections, the share of \emph{screw with hands} and \emph{screw with screwdriver} segments predicted as the other action drops from 6.6\% with the video expert to 1.6\% with Video + objects. Late fusion improves over the best single expert by 4.1 points in both settings: the video expert with detections and the object expert with annotated boxes. With detections, the object expert alone has a lower Top-1 than the video expert (37.8\% against 44.9\%), but a similar Top-5 (74.7\% against 74.2\%), so the objects narrow the candidate actions without deciding among them. With detections, only the object expert is correct in 12.4\% of the test segments and only the video expert in 19.5\%. With annotated boxes, only the object expert is correct in 30.6\%, which explains why late fusion outperforms the learned fusions most in this setting.

The learned fusions do not exploit this complementarity. SlowFast was trained on the same training split and its head reaches 99.1\% Top-1 on it, so a joint classifier fits the training labels from the video features alone and ignores the objects.

The Top-1 difference with detections is small (0.7 points over the best learned fusion) and not conclusive, but late fusion improves the Top-5 of this learned fusion by 2.0 points. The weight $\alpha$ chosen on the validation set increases from 0.40--0.60 with annotated boxes to 0.65--0.75 with detections, so the method relies more on the video expert when the objects come from the detector. Self-attention improves the object expert alone by 1.4 points over mean pooling with detections, against 1.0 with annotated boxes, possibly because attention lets each object weigh the others when some of them are non-active.

\begin{figure*}[t]
  \centering
  \setlength{\tabcolsep}{2pt}
  \begin{tabular}{@{}ccc@{}}
    \includegraphics[width=0.325\textwidth]{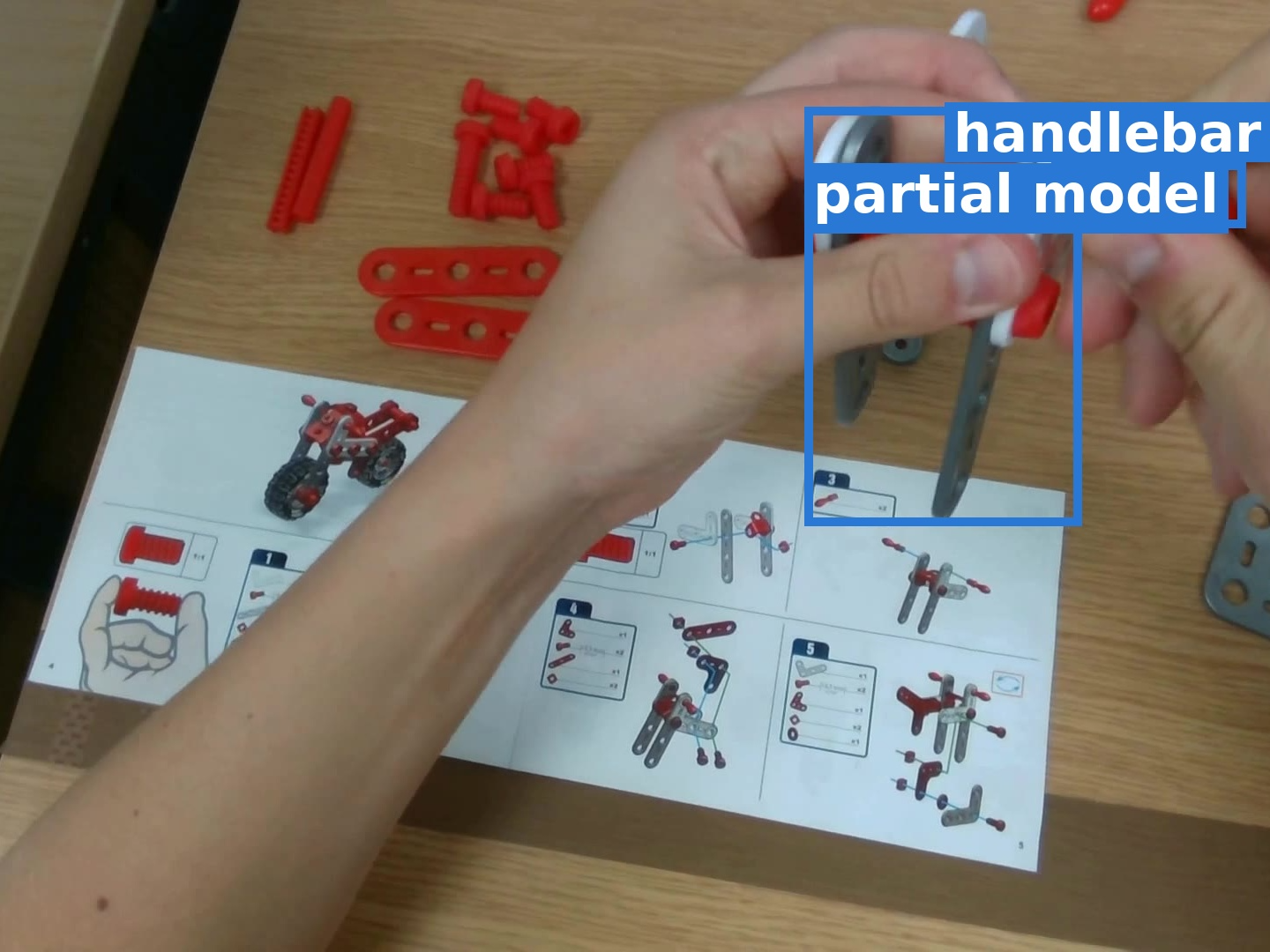} &
    \includegraphics[width=0.325\textwidth]{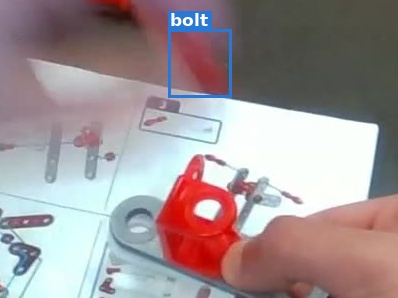} &
    \includegraphics[width=0.325\textwidth]{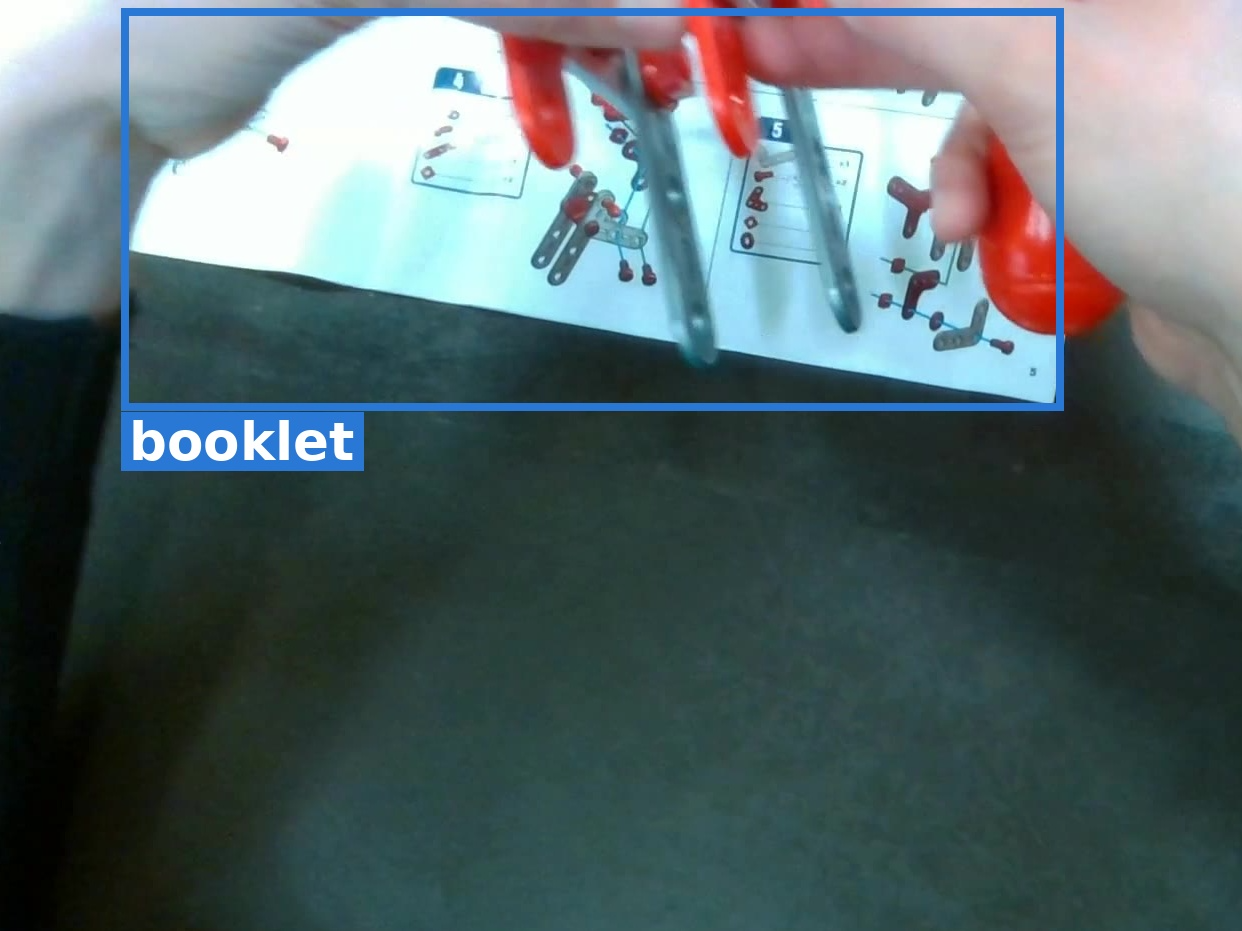} \\
    \includegraphics[width=0.325\textwidth]{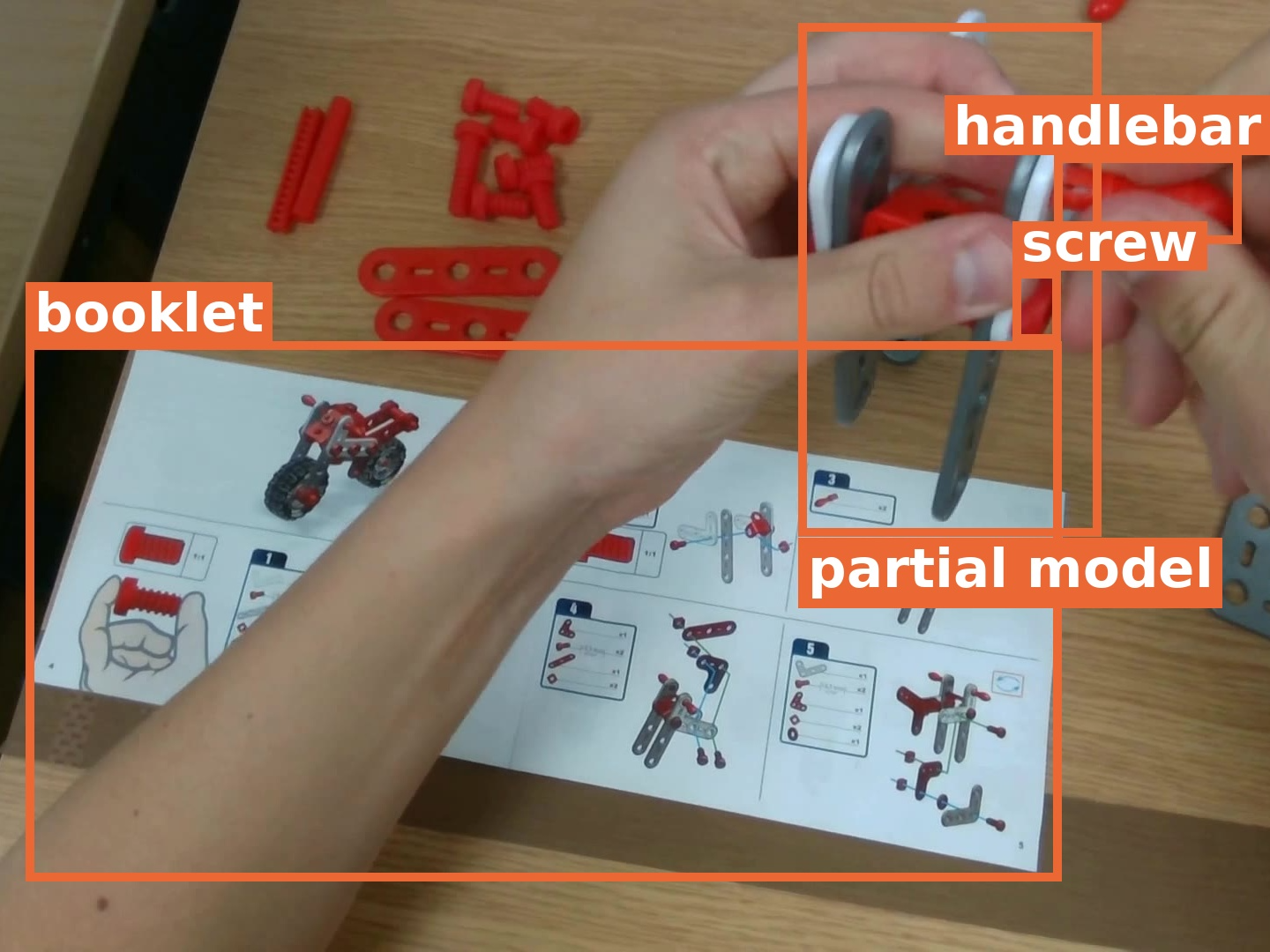} &
    \includegraphics[width=0.325\textwidth]{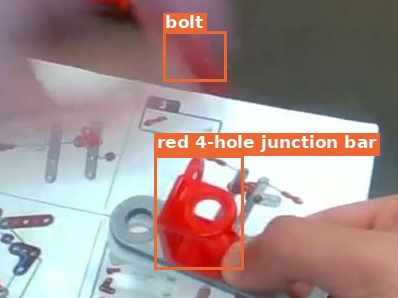} &
    \includegraphics[width=0.325\textwidth]{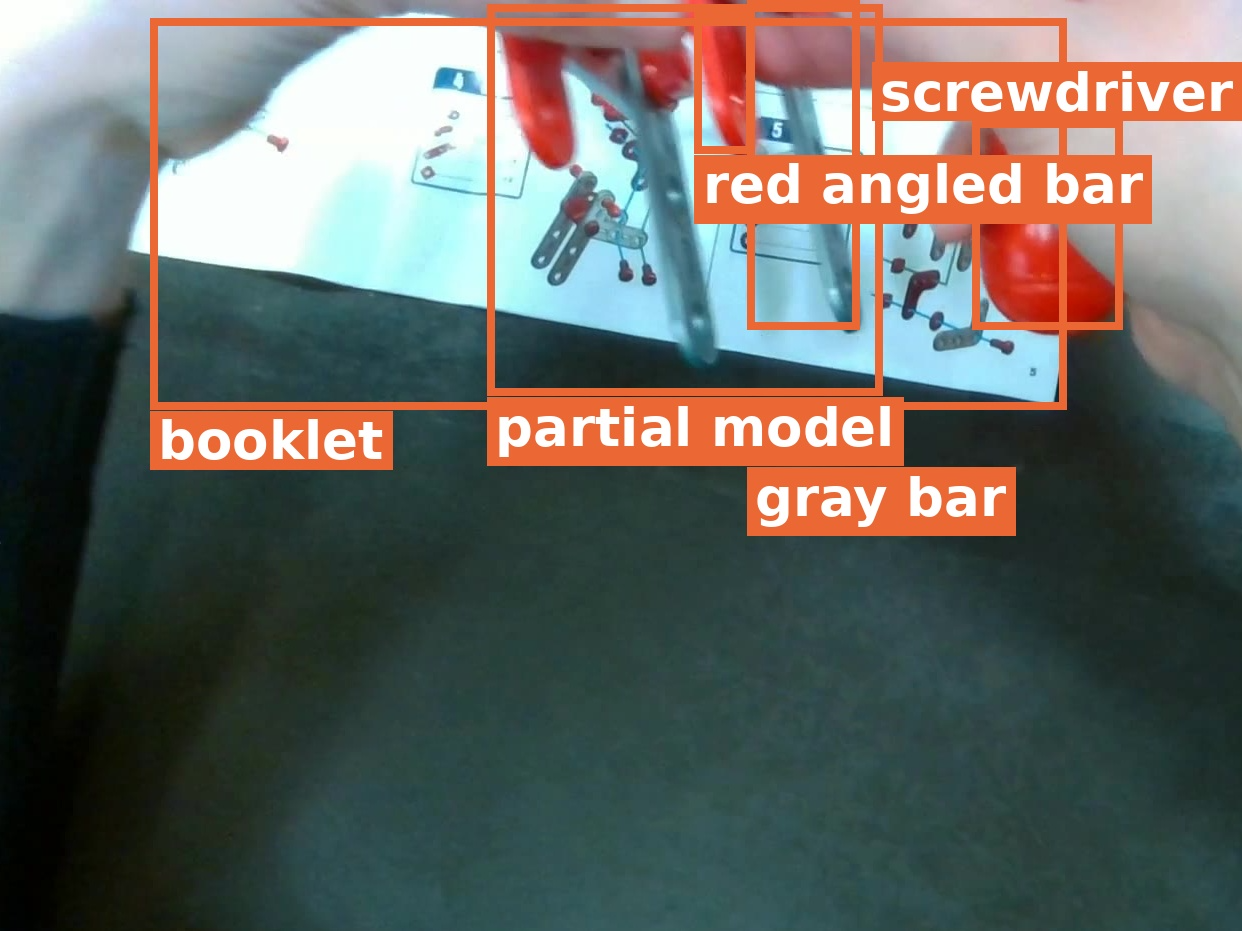} \\
    \includegraphics[width=0.325\textwidth]{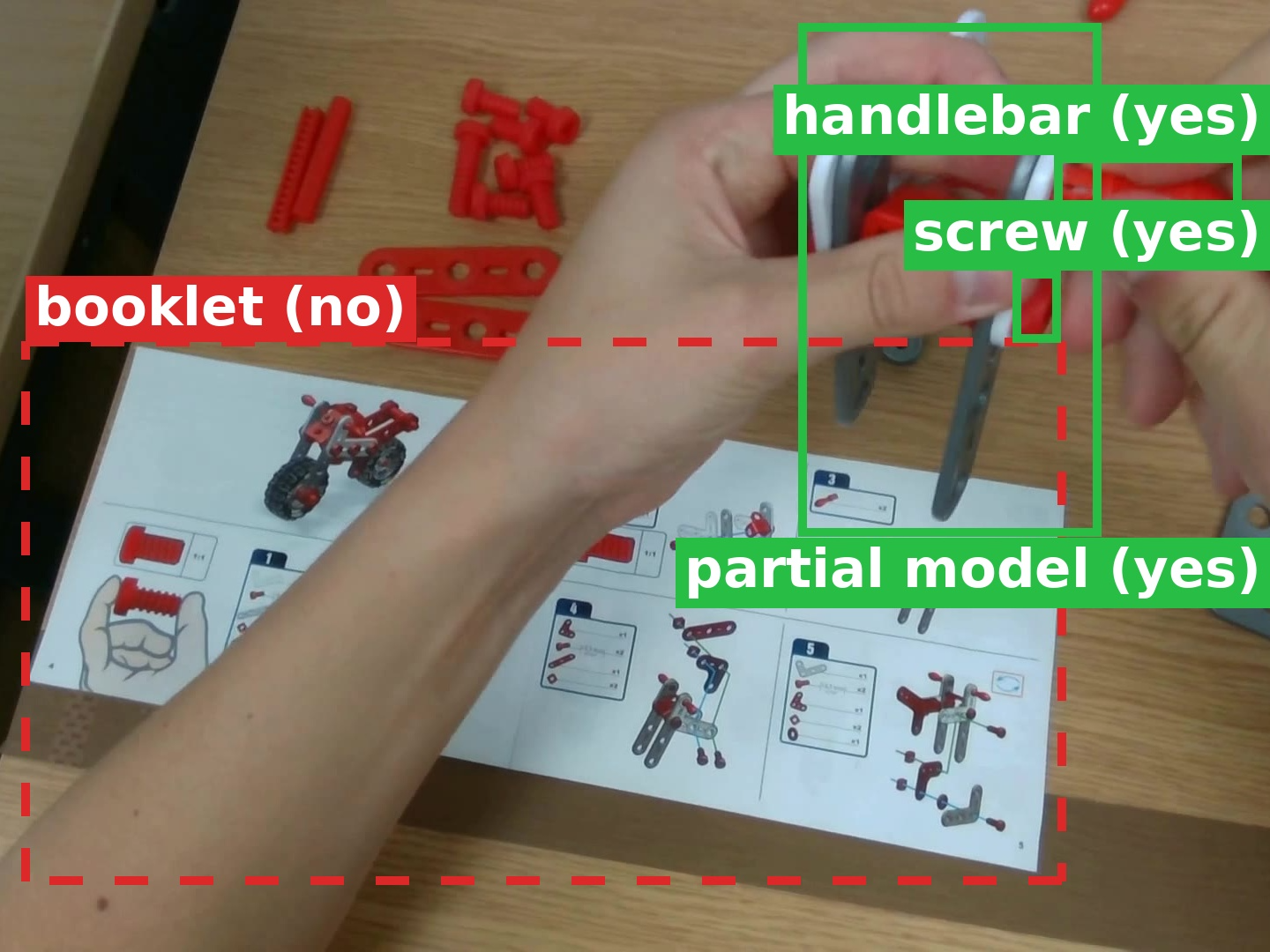} &
    \includegraphics[width=0.325\textwidth]{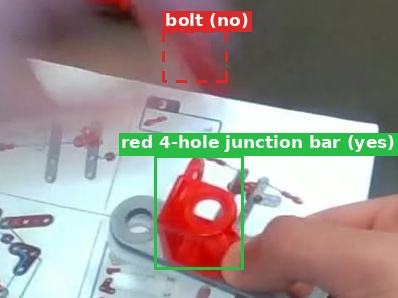} &
    \includegraphics[width=0.325\textwidth]{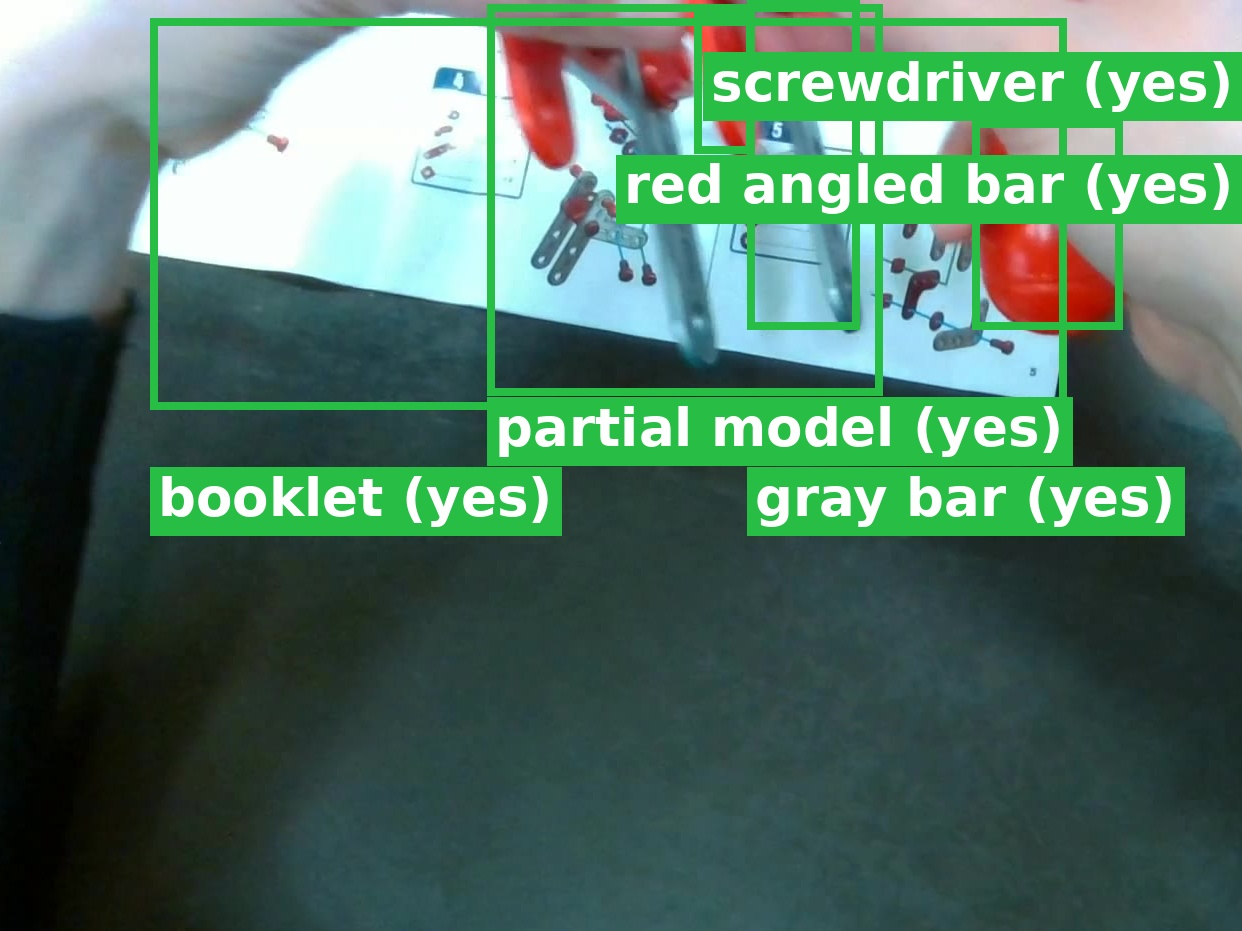} \\
    \parbox[t]{0.325\textwidth}{\scriptsize\centering
      (a) Ground truth: \emph{plug handlebar}\\
      Video expert: screw with hands \textcolor{red!75!black}{$\times$}\\
      Video + objects: screw with hands \textcolor{red!75!black}{$\times$}\\
      Ours: plug handlebar \textcolor{green!50!black}{\checkmark}} &
    \parbox[t]{0.325\textwidth}{\scriptsize\centering
      (b) Ground truth: \emph{take bolt}\\
      Video expert: take bolt \textcolor{green!50!black}{\checkmark}\\
      Video + objects: take bolt \textcolor{green!50!black}{\checkmark}\\
      Ours: align objects \textcolor{red!75!black}{$\times$}} &
    \parbox[t]{0.325\textwidth}{\scriptsize\centering
      (c) Ground truth: \emph{check booklet}\\
      Video expert: align wrench to bolt \textcolor{red!75!black}{$\times$}\\
      Video + objects: screw with screwdriver \textcolor{red!75!black}{$\times$}\\
      Ours: screw with screwdriver \textcolor{red!75!black}{$\times$}} \\
  \end{tabular}
  \caption{Qualitative results on the test set. Top: annotated boxes (blue). Middle: all detections (orange). Bottom: detections after the interaction filter (green: kept; red, dashed: removed), with the VLM's answer in parentheses. Predictions are the majority over five seeds. (a)~The filter removes the idle booklet, and our method corrects the prediction. (b)~The VLM answers \emph{no} for the bolt being taken, and the filter removes the only active object. (c)~The VLM answers \emph{yes} for every detection near the hands, and the non-active screwdriver remains.}
  \label{fig:qual}
\end{figure*}

Table~\ref{tab:frames} compares the frame selection strategies. The first row gives the Top-1 of maximize transitions with $K = 4$, and the other rows give the difference to it. \emph{Fixed offsets} takes raw frames at fixed offsets from the segment start, and \emph{uniform annotated} takes $K$ annotated frames equally spaced in time. \emph{High variability} and \emph{progressive discovery} select annotated frames where the set of object classes changes or new classes appear. With annotated boxes, the strategy changes Top-1 by up to 7.5 points, but with detections, all strategies stay within 1.2 points. More than four frames do not help with detections, likely because each additional frame adds non-active detections. \emph{Uniform annotated} is the best strategy on the test set, but the validation set selects \emph{maximize transitions}, and the method keeps this choice.

\subsection{Qualitative Results}\label{sec:exp_qual}

Fig.~\ref{fig:qual} shows three test segments with the annotated boxes, the detections, and the detections after the interaction filter. In Fig.~\ref{fig:qual}a, the filter removes the booklet lying on the table, and our method corrects the prediction of Video + objects. Figs.~\ref{fig:qual}b and~\ref{fig:qual}c show the two failure modes of the filter. In Fig.~\ref{fig:qual}b, the frame is blurred. The VLM answers \emph{no} for the bolt that the person is taking, so the filter removes the only active object. In Fig.~\ref{fig:qual}c, the VLM answers \emph{yes} for all five detections, and both Video + objects and our method predict \emph{screw with screwdriver}, the action of the non-active screwdriver. The second failure is the most frequent. The filter removes only 12.1\% of the non-active detections, compared with 3.7\% of the active ones, so its recall is the method's main limitation.

\section{CONCLUSIONS}\label{sec:conclusions}

This paper presents a method for egocentric action recognition that fuses a frozen SlowFast video expert with an attention-based object expert. Each object token includes zero-shot affordances generated by a VLM, and the same VLM provides an interaction affordance that removes detections that no hand manipulates. On MECCANO with the official active-object detector, the method improves the video expert and outperforms SlowFast and object-aware RGB baselines, and the affordances and the interaction filter give small but consistent gains. The analysis of the drop from annotated boxes to detections shows that the main obstacle for object-based recognition with a real detector is non-active detections, a problem that evaluation with annotated boxes hides. These results suggest that object-based methods should also be evaluated with detections.

Future work will use the temporal context of the segment, rather than a single frame, to decide whether an object is manipulated. It will also test hand-object detectors that localize only objects in contact with the hands, and evaluate our method on other egocentric datasets.



{\small
\bibliographystyle{ieee}
\bibliography{egbib}

\begin{thebibliography}{10}\itemsep=-1pt

\bibitem{bahl2023affordances}
S.~Bahl, R.~Mendonca, L.~Chen, U.~Jain, and D.~Pathak.
\newblock Affordances from human videos as a versatile representation for robotics.
\newblock In {\em IEEE/CVF Conference on Computer Vision and Pattern Recognition (CVPR)}, pages 13778--13790, 2023.

\bibitem{bertasius2021space}
G.~Bertasius, H.~Wang, and L.~Torresani.
\newblock Is space-time attention all you need for video understanding?
\newblock In {\em International Conference on Machine Learning (ICML)}, volume 139 of {\em Proceedings of Machine Learning Research}, pages 813--824, 2021.

\bibitem{damen2022rescaling}
D.~Damen, H.~Doughty, G.~M. Farinella, A.~Furnari, E.~Kazakos, J.~Ma, D.~Moltisanti, J.~Munro, T.~Perrett, W.~Price, and M.~Wray.
\newblock Rescaling egocentric vision: Collection, pipeline and challenges for {EPIC-KITCHENS-100}.
\newblock {\em International Journal of Computer Vision}, 130:33--55, 2022.

\bibitem{dosovitskiy2021image}
A.~Dosovitskiy, L.~Beyer, A.~Kolesnikov, D.~Weissenborn, X.~Zhai, T.~Unterthiner, M.~Dehghani, M.~Minderer, G.~Heigold, S.~Gelly, J.~Uszkoreit, and N.~Houlsby.
\newblock An image is worth 16x16 words: Transformers for image recognition at scale.
\newblock In {\em International Conference on Learning Representations (ICLR)}, 2021.

\bibitem{engelbracht2025spotlight}
T.~Engelbracht, R.~Zurbr{\"u}gg, M.~Pollefeys, H.~Blum, and Z.~Bauer.
\newblock {SpotLight}: Robotic scene understanding through interaction and affordance detection.
\newblock In {\em IEEE-RAS International Conference on Humanoid Robots (Humanoids)}, 2025.

\bibitem{fathi2011learning}
A.~Fathi, X.~Ren, and J.~M. Rehg.
\newblock Learning to recognize objects in egocentric activities.
\newblock In {\em IEEE Conference on Computer Vision and Pattern Recognition (CVPR)}, pages 3281--3288, 2011.

\bibitem{feichtenhofer2019slowfast}
C.~Feichtenhofer, H.~Fan, J.~Malik, and K.~He.
\newblock {SlowFast} networks for video recognition.
\newblock In {\em IEEE/CVF International Conference on Computer Vision (ICCV)}, pages 6202--6211, 2019.

\bibitem{gibson1979ecological}
J.~J. Gibson.
\newblock {\em The Ecological Approach to Visual Perception}.
\newblock Houghton Mifflin, Boston, 1979.

\bibitem{gemini}
{Google DeepMind}.
\newblock {Gemini 3.1 Flash-Lite} model card.
\newblock \url{https://deepmind.google/models/model-cards/gemini-3-1-flash-lite/}, 2026.

\bibitem{grauman2022ego4d}
K.~Grauman, A.~Westbury, E.~Byrne, Z.~Chavis, A.~Furnari, R.~Girdhar, et~al.
\newblock {Ego4D}: Around the world in 3,000 hours of egocentric video.
\newblock In {\em IEEE/CVF Conference on Computer Vision and Pattern Recognition (CVPR)}, pages 18995--19012, 2022.

\bibitem{haralovic2026g3ego}
M.~Haralovi{\'c}, A.~Ramakrishnan, and E.~Talavera~Martinez.
\newblock {G3Ego}: Gaze-guided graphs for egocentric action understanding.
\newblock arXiv:2608.20157 (CONTEXTUS Workshop, ECCV), 2026.

\bibitem{kazakos2019epic}
E.~Kazakos, A.~Nagrani, A.~Zisserman, and D.~Damen.
\newblock {EPIC-Fusion}: Audio-visual temporal binding for egocentric action recognition.
\newblock In {\em IEEE/CVF International Conference on Computer Vision (ICCV)}, pages 5492--5501, 2019.

\bibitem{kini2023egocentric}
J.~Kini, S.~Fleischer, I.~Dave, and M.~Shah.
\newblock Egocentric {RGB+Depth} action recognition in industry-like settings.
\newblock arXiv:2309.13962, 2023.

\bibitem{li2018eye}
Y.~Li, M.~Liu, and J.~M. Rehg.
\newblock In the eye of beholder: Joint learning of gaze and actions in first person video.
\newblock In {\em European Conference on Computer Vision (ECCV)}, pages 619--635, 2018.

\bibitem{lin2022egocentric}
K.~Q. Lin, J.~Wang, M.~Soldan, M.~Wray, R.~Yan, E.~Z. Xu, D.~Gao, R.-C. Tu, W.~Zhao, W.~Kong, C.~Cai, H.~Wang, D.~Damen, B.~Ghanem, W.~Liu, and M.~Z. Shou.
\newblock Egocentric video-language pretraining.
\newblock In {\em Advances in Neural Information Processing Systems (NeurIPS)}, volume~35, 2022.

\bibitem{materzynska2020something}
J.~Materzynska, T.~Xiao, R.~Herzig, H.~Xu, X.~Wang, and T.~Darrell.
\newblock {Something-Else}: Compositional action recognition with spatial-temporal interaction networks.
\newblock In {\em IEEE/CVF Conference on Computer Vision and Pattern Recognition (CVPR)}, pages 1049--1059, 2020.

\bibitem{murlabadia2024aff}
L.~Mur-Labadia, R.~Martinez-Cantin, J.~J. Guerrero, G.~M. Farinella, and A.~Furnari.
\newblock {AFF-ttention}! {Affordances} and attention models for short-term object interaction anticipation.
\newblock In {\em European Conference on Computer Vision (ECCV)}, pages 167--184, 2024.

\bibitem{nagarajan2019grounded}
T.~Nagarajan, C.~Feichtenhofer, and K.~Grauman.
\newblock Grounded human-object interaction hotspots from video.
\newblock In {\em IEEE/CVF International Conference on Computer Vision (ICCV)}, pages 8688--8697, 2019.

\bibitem{nagarajan2020ego}
T.~Nagarajan, Y.~Li, C.~Feichtenhofer, and K.~Grauman.
\newblock {Ego-Topo}: Environment affordances from egocentric video.
\newblock In {\em IEEE/CVF Conference on Computer Vision and Pattern Recognition (CVPR)}, pages 163--172, 2020.

\bibitem{pirsiavash2012detecting}
H.~Pirsiavash and D.~Ramanan.
\newblock Detecting activities of daily living in first-person camera views.
\newblock In {\em IEEE Conference on Computer Vision and Pattern Recognition (CVPR)}, pages 2847--2854, 2012.

\bibitem{qian2024affordancellm}
S.~Qian, W.~Chen, M.~Bai, X.~Zhou, Z.~Tu, and L.~E. Li.
\newblock {AffordanceLLM}: Grounding affordance from vision language models.
\newblock In {\em IEEE/CVF Conference on Computer Vision and Pattern Recognition Workshops (CVPRW)}, pages 7587--7597, 2024.

\bibitem{ragusa2023meccano}
F.~Ragusa, A.~Furnari, and G.~M. Farinella.
\newblock {MECCANO}: A multimodal egocentric dataset for humans behavior understanding in the industrial-like domain.
\newblock {\em Computer Vision and Image Understanding}, 235:103764, 2023.

\bibitem{ragusa2021meccano}
F.~Ragusa, A.~Furnari, S.~Livatino, and G.~M. Farinella.
\newblock The {MECCANO} dataset: Understanding human-object interactions from egocentric videos in an industrial-like domain.
\newblock In {\em IEEE/CVF Winter Conference on Applications of Computer Vision (WACV)}, pages 1569--1578, 2021.

\bibitem{ren2015faster}
S.~Ren, K.~He, R.~Girshick, and J.~Sun.
\newblock Faster {R-CNN}: Towards real-time object detection with region proposal networks.
\newblock In {\em Advances in Neural Information Processing Systems (NeurIPS)}, volume~28, 2015.

\bibitem{sanh2019distilbert}
V.~Sanh, L.~Debut, J.~Chaumond, and T.~Wolf.
\newblock {DistilBERT}, a distilled version of {BERT}: Smaller, faster, cheaper and lighter.
\newblock arXiv:1910.01108 (EMC$^2$ Workshop, NeurIPS), 2019.

\bibitem{sener2022assembly101}
F.~Sener, D.~Chatterjee, D.~Shelepov, K.~He, D.~Singhania, R.~Wang, and A.~Yao.
\newblock {Assembly101}: A large-scale multi-view video dataset for understanding procedural activities.
\newblock In {\em IEEE/CVF Conference on Computer Vision and Pattern Recognition (CVPR)}, pages 21096--21106, 2022.

\bibitem{shan2020understanding}
D.~Shan, J.~Geng, M.~Shu, and D.~F. Fouhey.
\newblock Understanding human hands in contact at internet scale.
\newblock In {\em IEEE/CVF Conference on Computer Vision and Pattern Recognition (CVPR)}, pages 9869--9878, 2020.

\bibitem{shiota2024egocentric}
T.~Shiota, M.~Takagi, K.~Kumagai, H.~Seshimo, and Y.~Aono.
\newblock Egocentric action recognition by capturing hand-object contact and object state.
\newblock In {\em IEEE/CVF Winter Conference on Applications of Computer Vision (WACV)}, pages 6541--6551, 2024.

\bibitem{tong2022videomae}
Z.~Tong, Y.~Song, J.~Wang, and L.~Wang.
\newblock {VideoMAE}: Masked autoencoders are data-efficient learners for self-supervised video pre-training.
\newblock In {\em Advances in Neural Information Processing Systems (NeurIPS)}, volume~35, 2022.

\bibitem{tonini2026towards}
F.~Tonini, A.~Conti, L.~Vaquero, C.~Beyan, and E.~Ricci.
\newblock Towards unconstrained human-object interaction.
\newblock In {\em IEEE International Conference on Automatic Face and Gesture Recognition (FG)}, 2026.

\bibitem{tran2015learning}
D.~Tran, L.~Bourdev, R.~Fergus, L.~Torresani, and M.~Paluri.
\newblock Learning spatiotemporal features with {3D} convolutional networks.
\newblock In {\em IEEE International Conference on Computer Vision (ICCV)}, pages 4489--4497, 2015.

\bibitem{vaswani2017attention}
A.~Vaswani, N.~Shazeer, N.~Parmar, J.~Uszkoreit, L.~Jones, A.~N. Gomez, {\L}.~Kaiser, and I.~Polosukhin.
\newblock Attention is all you need.
\newblock In {\em Advances in Neural Information Processing Systems (NeurIPS)}, volume~30, 2017.

\bibitem{wang2025actionclip}
M.~Wang, J.~Xing, J.~Mei, Y.~Liu, and Y.~Jiang.
\newblock {ActionCLIP}: Adapting language-image pretrained models for video action recognition.
\newblock {\em IEEE Transactions on Neural Networks and Learning Systems}, 36:625--637, 2025.

\bibitem{zhao2023learning}
Y.~Zhao, I.~Misra, P.~Kr{\"a}henb{\"u}hl, and R.~Girdhar.
\newblock Learning video representations from large language models.
\newblock In {\em Proceedings of the IEEE/CVF Conference on Computer Vision and Pattern Recognition (CVPR)}, pages 6586--6597, 2023.

\end{thebibliography}
}

\end{document}